\documentclass[10pt,twocolumn,letterpaper]{article}

\usepackage[pagenumbers]{cvpr} 

\usepackage{booktabs}
\usepackage{colortbl}
\usepackage{listings}
\usepackage{multirow}
\usepackage{makecell, cellspace, caption}
\usepackage[dvipsnames]{xcolor}

\definecolor{Gray}{gray}{0.92}
\newcolumntype{a}{>{\columncolor{Gray}}r}
\definecolor{Gray2}{gray}{0.72}
\newcolumntype{g}{>{\columncolor{Gray2}}r}

\definecolor{codegreen}{rgb}{0,0.6,0}
\definecolor{codegray}{rgb}{0.5,0.5,0.5}
\definecolor{codepurple}{rgb}{0.58,0,0.82}
\definecolor{backcolour}{rgb}{0.95,0.95,0.92}

\lstdefinestyle{mystyle}{
    backgroundcolor=\color{backcolour},   
    commentstyle=\color{codegreen},
    keywordstyle=\color{magenta},
    numberstyle=\tiny\color{codegray},
    stringstyle=\color{codepurple},
    basicstyle=\ttfamily\footnotesize,
    breakatwhitespace=false,         
    breaklines=true,                 
    captionpos=b,                    
    keepspaces=true,                 
    numbers=left,                    
    numbersep=5pt,                  
    showspaces=false,                
    showstringspaces=false,
    showtabs=false,                  
    tabsize=2
}

\definecolor{cvprblue}{rgb}{0.21,0.49,0.74}
\usepackage[pagebackref,breaklinks,colorlinks,citecolor=cvprblue]{hyperref}

\providecommand{\sr}{\mbox{\sc{Super-Res}}\xspace}
\providecommand{\metric}{\mbox{\sc{CLIPScore}}\xspace}
\providecommand{\dataset}{\mbox{\sc{S2-NAIP}}\xspace}

\def\paperID{951} 
\def\confName{CVPR}
\def\confYear{2024}

\begin{document}

\title{\emph{Zooming Out on Zooming In:} Advancing Super-Resolution for Remote Sensing}

\author{Piper Wolters \quad Favyen Bastani \quad Aniruddha Kembhavi \\
Allen Institute for AI \\
{\tt\small \{piperw,favyenb,anik\}@allenai.org}
}


\twocolumn[{%
\renewcommand\twocolumn[1][]{#1}%
\maketitle
\begin{center}
    \centering
    \captionsetup{type=figure}
    \includegraphics[width=\linewidth]{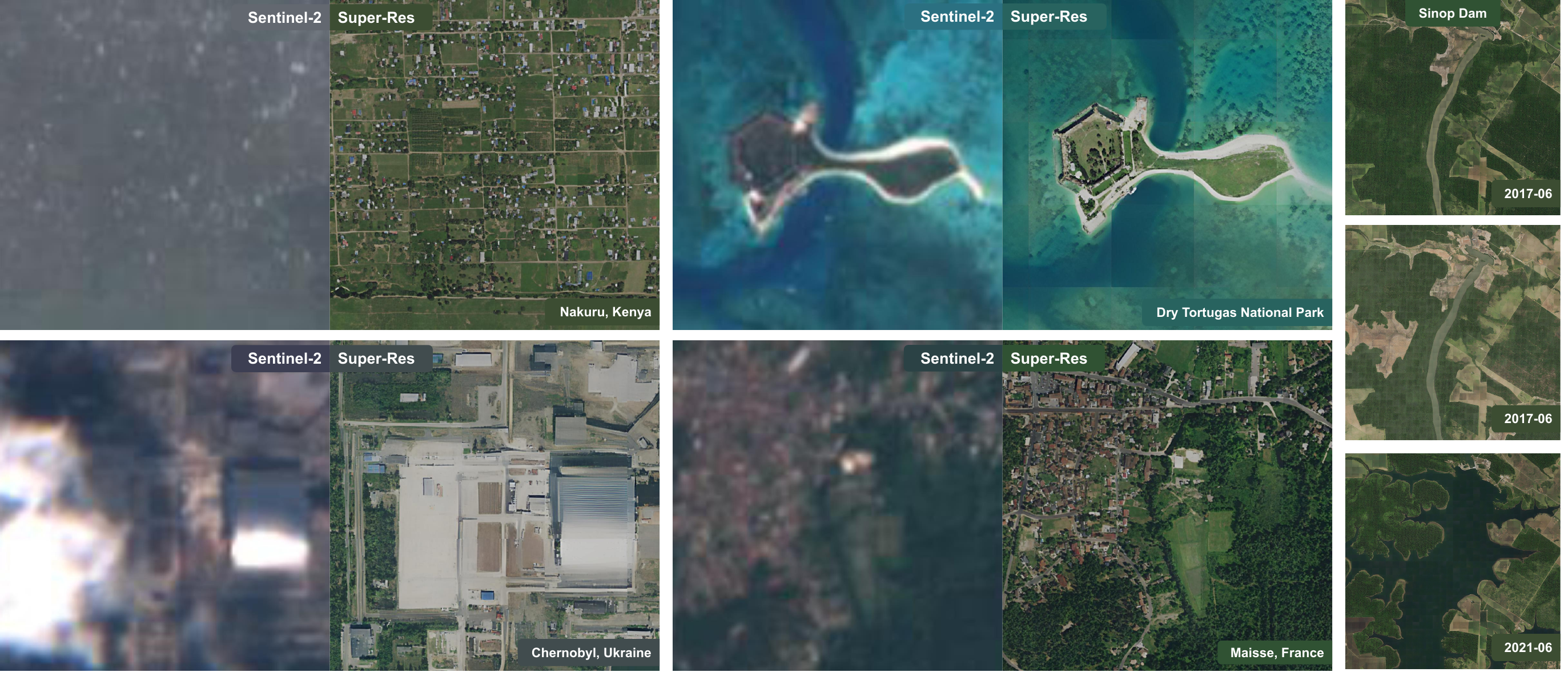}
    \captionof{figure}{Display of the incredible power of super-resolution for remote sensing imagery. High-resolution satellite imagery is not available for free worldwide, and a public source such as NAIP is restricted to every 2-3 years in the US. On the other hand, Sentinel-2 imagery is global, free, and has a revisit rate of 5-10 days, so with super-resolution methods, we can generate high-resolution imagery globally and frequently, especially in places that have disproportionately less public imagery.}
    \label{fig:teaser}
\end{center}%
}]

\begin{abstract}
\vspace{-0.3cm}
Super-Resolution for remote sensing has the potential for huge impact on planet monitoring by producing accurate and realistic high resolution imagery on a frequent basis and a global scale. Despite a lot of attention, several inconsistencies and challenges have prevented it from being deployed in practice. These include the lack of effective metrics, fragmented and relatively small-scale datasets for training, insufficient comparisons across a suite of methods, and unclear evidence for the use of super-resolution outputs for machine consumption. This work presents a new metric for super-resolution, \metric, that corresponds far better with human judgments than previous metrics on an extensive study. We use \metric to evaluate four standard methods on a new large-scale dataset, \dataset, and three existing benchmark datasets, and find that generative adversarial networks easily outperform more traditional L2 loss-based models and are more semantically accurate than modern diffusion models. We also find that using \metric as an auxiliary loss can speed up the training of GANs by 18x and lead to improved outputs, resulting in an effective model in diverse geographies across the world which we will release publicly. The dataset, pre-trained model weights, and code are available at \href{https://github.com/allenai/satlas-super-resolution/tree/main}{this URL}.

\end{abstract}

\vspace{-0.5cm}

\section{Introduction}

High-resolution satellite and aerial imagery have the potential to change the landscape of environmental and climate monitoring applications. Images from these sensors provide the ability to count individual trees \cite{treecounting}, classify crop types and conditions \cite{finecrops}, map out detailed land use categories \cite{lulc} and track glacial conditions \cite{glaciers}. Low resolution imagery limits these observations to coarse designations of regions such as forest, farmland, residential or ice. Public high-resolution imagery is available infrequently in a few developed countries and commercial global imagery is expensive and can be five or more years out of date. With the bounded spatio-temporal coverage of public domain high-resolution imagery and cost-prohibitive commercial imagery, many use cases cannot be scaled up globally. 

This lack of free high-resolution imagery has prompted research in \textbf{Super-Resolution for Remote Sensing}, where computer vision methods are used to upsample low-resolution imagery. Past works include spline models \cite{spline}, convolutional neural networks \cite{highresnet, cnnucmcoffee} and generative networks \cite{rssrgan, diffusion}. Within this field are diverse datasets with varying satellites, spatial resolutions \cite{mus2,probav} and earth coverage \cite{massbuild,probav}. 
Despite this attention and potential for impact, Super-Resolution (\sr) for remote sensing faces many challenges that are limiting progress.

First, metrics for \sr have serious and well known issues, see Figure \ref{fig:blur}. It is crucial to establish a metric that closely aligns with human perception and reflects the accuracy and realism of \sr outputs. We run an extensive human evaluation study to assess the quality of \sr outputs, and then compare human preferences to those of multiple pixel-wise, perceptual, and model-based metrics. We determine that model-based metrics easily outperform pixel-wise ones, and that using modern and powerful visual encoders is most effective. Based on this analysis and inspired by \cite{clipscore}, we propose a new metric for this domain, \metric.

Second, existing remote sensing datasets are relatively small in comparison to those for natural images (see Table \ref{tab:datasets}), and often imagery is sourced from commercial vendors, making it expensive to deploy models in practice. We present a new large-scale public domain remote sensing \sr dataset, covering 113K $km^2$ using free imagery from Sentinel-2 \cite{sentinel} and the US National Agriculture Imagery Program (NAIP). There are at least 18 Sentinel-2 images for each NAIP image (to support multi-view super-resolution), along with geographic and temporal metadata. This large amount of free imagery allows us to determine that scaling up indeed increases performance, and models can be continually updated with the reoccurring public data for years to come.

Third, while many diverse models have been proposed, rarely are models compared side-by-side, especially models that are biased towards the simple and shallow metrics PSNR and SSIM \cite{highresnet, ram, fusion, multiscaleresnet} with models that are more geared towards perceptual quality \cite{attngan, diffusion, oli2msi, diffusionucm}. We perform comprehensive experiments on four remote sensing \sr datasets with two L2 loss-based methods, a GAN, and a diffusion model. We find that the L2 loss-based methods lag behind by large margins and that the GAN outperforms other methods in terms of our newly proposed metric, \metric (and also the popular LPIPS metric) on all four datasets. We also find that training GANs with a CLIPScore based loss speeds up training dramatically (18x speedup) and further improves performance. With the increasing emphasis on avoiding hallucinations by generative methods, we run a human study to determine the accuracy of human made structures and find that the GAN is superior to the diffusion model. This body of evidence suggests that despite GANs falling out of favor to modern diffusion models, particularly in the natural image domain, they nevertheless perform really well and provide much faster inference for remote sensing \sr.

Fourth, while there is some work exploring the use of \sr images as input to downstream tasks \cite{semantically} in lieu of the original low-resolution images, the evidence of improvement over directly inputting the low-resolution images is unclear. And although representation learning has received much interest in computer vision, the use of \sr as a representation learning mechanism is under-explored. We conduct a set of experiments to address both of these directions and find that \sr is a powerful representation learning mechanism that shows promise on downstream remote sensing tasks; but also find that training models on \sr outputs may not yet be more effective than training on the original low-resolution images for these tasks. Based on these findings, we establish that although \sr outputs can have a large impact for planet monitoring, they are primarily effective for human consumption (visualization); more progress is required for them to be effective for machine consumption.

We leverage our findings to build an effective model for \sr that works well in diverse geographies. Furthermore, we deploy the model globally to regularly compute up-to-date high-resolution imagery that is freely available. All data, code, models, and pretrained weights are available at \href{https://github.com/allenai/satlas-super-resolution/tree/main}{this URL}.

In summary, our contributions are:
\begin{enumerate}[noitemsep]
    \item \metric, a new metric for \sr that strongly corresponds with human preferences.
    \item \dataset, a large-scale, public-domain dataset for remote-sensing \sr.
    \item An extensive evaluation of \sr methods on four remote sensing datasets, and a determination that GANs are the current state-of-the-art.
    \item Demonstrating that transferring \sr weights to downstream tasks works well, but machine consumption of \sr outputs is ineffective.
    \item A new global model for \sr that leverages the above contributions.
\end{enumerate}

\vspace{-0.1cm}
\section{Related Work}

\sr has a rich history in computer vision \cite{srcnn,edsr,rdn,realesrgan}, with recent methods achieving very impressive results \cite{gigagan,resshift}. This paper will focus on \sr for remote sensing, hence we provide context for this domain.

\textbf{Remote Sensing Super-Resolution Metrics}. The two most widely used metrics in \sr are Peak Signal to Noise Ratio (PSNR) and Structural Similarity Index Measure (SSIM) \cite{ssim}. PSNR is a pixel-based metric that is the inverse of an L2 loss; and SSIM was proposed as a perceptual metric based on image qualities like luminance and contrast. They are both simple functions, as briefly addressed by \cite{Beaulieu2018DeepIT,mus2}, that are unable to handle ambiguities like shadow direction or nadir angle and physical changes like crop harvest cycles. Figure \ref{fig:blur} shows how these are not influenced by crucial perceptual nuances such as blur. Märtens et al. \cite{probav} propose cPSNR, a variation of PSNR that handles misalignment and brightness differences but the lack of contextual information remains an issue.

Drawbacks with standard metrics have prompted perceptual metrics like Learned Perceptual Image Patch Similarity (LPIPS) \cite{lpips}, which is proven to align with human judgement and is adopted in the natural image domain \cite{lpips2,lpips3,lpips4}. Still, much of the remote sensing community only reports SSIM or pixel-wise metrics \cite{s2planetsr,ram,deepsum}, including two of the most recently proposed super-resolution benchmarks, SEN2VENµS and WorldStrat \cite{sen2venus,worldstrat}. We take inspiration from CLIPScore \cite{clipscore} to find a new metric for \sr.

\textbf{Remote Sensing Super-Resolution Datasets}. Existing datasets can be categorized based on the satellite sensors used, whether they are targeting single image \sr \cite{oli2msi, sen2venus} or multi-image \sr \cite{probav, worldstrat, mus2}, and the scale of the data, both in $km^2$ covered and number of pixels. 

Many datasets curate pairs of low-resolution and high-resolution images, from one \cite{probav} or multiple \cite{worldstrat, multiscaleresnet, oli2msi, sen2venus} sensors. In multi-sensor datasets, many utilize Sentinel-2 as the low-resolution image source \cite{worldstrat, sen2venus, mus2}, likely because it is free and maintains a good balance of spatial resolution and temporal revisits. Other works obtain imagery from sources such as Google Earth \cite{rrsgan}, PlanetScope \cite{planetscope}, PeruSAT \cite{perusat}, China GF 1 \cite{chinagf1} and WorldView \cite{mus2}. 

Even with the many datasets that are intended to be standard benchmarks for remote sensing \sr \cite{mus2, worldstrat, sen2venus, oli2msi}, there are still papers that evaluate their methods on piecemeal datasets \cite{evalsrgan, attngan, perusat}. Also, synthetic datasets are often generated by bicubicly downsampling high-resolution images to create artificial pairs \cite{srfeinr, tesagan, diffusionucm}, which is unrealistic and biased by standard metrics.

Of the proposed benchmark datasets, none of them are large-scale nor are they entirely obtained from free sources. In this paper, we present \dataset, a large-scale dataset with over 1.2 million pairs of public domain images, covering 113 thousand $km^2$. Table \ref{tab:datasets} shows information about existing benchmarks in comparison to \dataset.

\begin{table}
    \centering
    \footnotesize
    \setlength\tabcolsep{4pt}
    \begin{tabular}{|l|r|r|r|r|r|r|r|}
        \hline
         & Sensors & Pairs & Input & HR Res. & $km^2$ \\
        \hline
        \textbf{S2-NAIP} & S2,NAIP & 1,200K & 18 & 512x512 & 113K  \\
        SEN2VENµS & S2,VENµS & 133K & 1 & 256x256 & 0.806K \\
        WorldStrat & S2,SPOT & 4K & 16 & 1054x1054 & 10K \\
        OLI2MSI & LandSat,S2 & 5.225K & 1 & 480x480 & 0.2K \\
        PROBA-V & PROBA-V & 2.368K & 19 & 384x384 & 3,000K \\
        MuS2 & S2,WV-2 & 0.091K & 14 & 1965x1686 & 2.5K \\
        \hline
    \end{tabular}
    \caption{Metadata about existing datasets: the low-resolution (LR) and high-resolution (HR) sensors used, the number of HR, LR pairs, the number of LR input images, the spatial resolution of HR (HR Res.) and coverage in terms of $km^2$.}
\label{tab:datasets}
\end{table}

\textbf{Remote Sensing Super-Resolution Methods}. 
Broadly, recent \sr methods fall into the categories of convolutional \cite{deepsum, ram, highresnet}, generative adversarial \cite{rssrgan, oli2msi, tesagan}, and diffusion models \cite{diffusion, enhancediffusion, diffusionucm}. There is a lack of evaluations across these three diverse method families, likely because it is difficult to compare them without strong and consistent metrics. We provide a comprehensive analysis on methods from each of these categories on four remote sensing \sr datasets, enabling us to determine the most effective method type for remote sensing \sr. A tangentially related line of work uses older images of a reference image as input to the model along with the low resolution image at the desired timestamp \cite{triplets}; this produces impressive results, but requires high resolution imagery at inference time, which limits its use in practice.

\vspace{-0.1cm}
\section{Metrics}
\begin{figure}[h!]
    \centering
    \includegraphics[width=\linewidth]{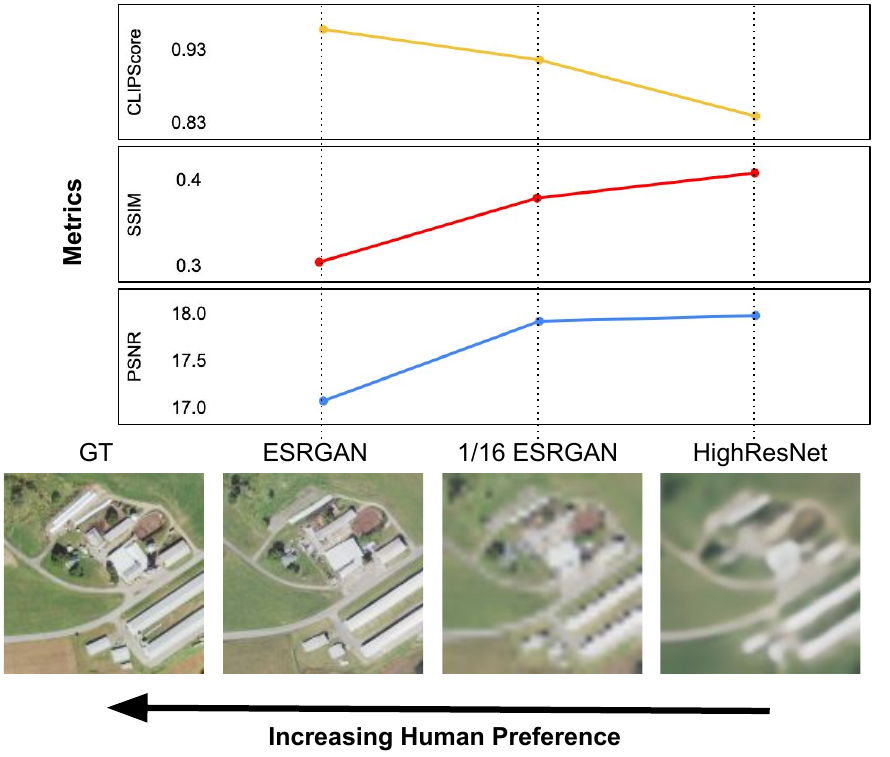}
    \caption{Example of a target image (GT), an ESRGAN output at full resolution as well as downsampled 16x, and a HighResNet output, with corresponding metrics. Note that the four images are ordered from best to worst based on human preference, and PSNR and SSIM increase in an opposite trend. Our proposed CLIPScore more closely matches human judgement.}
    \label{fig:blur}
\end{figure}

\vspace{-0.1cm}
PSNR and SSIM are the most commonly used metrics to evaluate \sr outputs for remote sensing. These metrics however, have many inherent issues. Previous work in the natural image domain has acknowledged these, including the well-known finding that blurring an image will cause significant perceptual change but little difference will be reflected in PSNR \cite{lpips}. Figure \ref{fig:blur} showcases this in the remote sensing domain. Although the two \sr outputs on the right are quite poor, with one being blurry and the other being sharply downsampled and disfiguring structures, they attain high PSNR and SSIM scores, far beyond an image output from an ESRGAN model which resembles the ground truth image far more closely to the human eye. We now present an extensive human evaluation study to evaluate a suite of metrics and present an image similarity metric based on CLIP \cite{clip}.

\subsection{Super-Resolution Human Judgement Dataset}

To find a metric that can effectively assess perceptual quality of remote sensing \sr outputs, we run a human evaluation study to gather a dataset of human judgements on our \dataset dataset and WorldStrat \cite{worldstrat}. Preferences are gathered using Amazon Mechanical Turk (AMT), where each AMT worker was asked to pick, between two model outputs, which one is closer to the target image. 

Pairs of outputs were chosen from a set of generated outputs from multiple models on each of 20,000 datapoints from \dataset and 381 datapoints from the WorldStrat. A total of 11,524 pair-wise annotations were collected. More details on this process are in supplementary Section A.1.

\begin{figure}
    \centering
    \includegraphics[width=\linewidth]{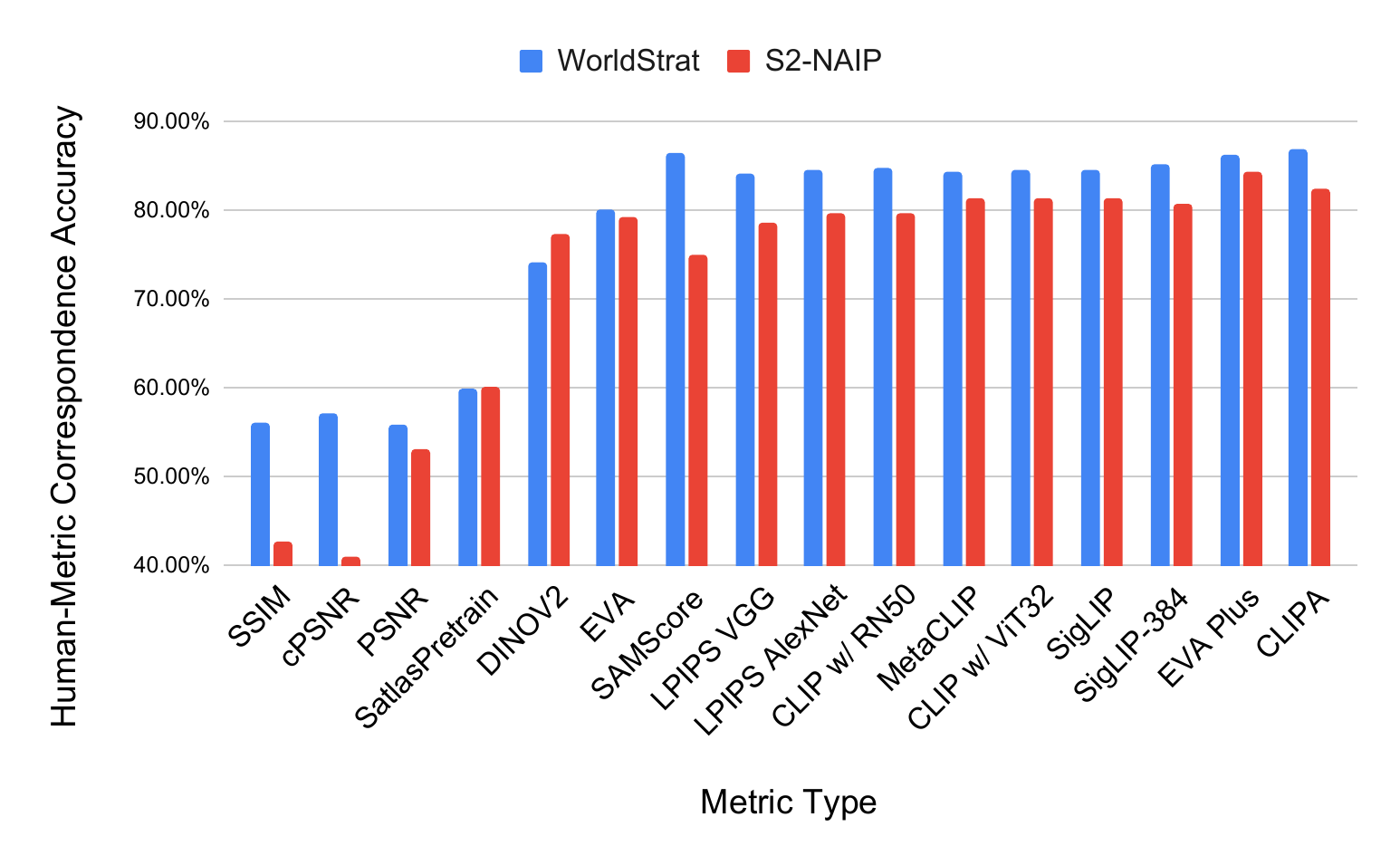}
    \caption{The level of accuracy between human preferences and those generated by the various metrics. The x-axis is ordered from worst to best average accuracy between the two datasets. The y-axis is adjusted to a range of 40\% to 90\% to better show the difference in accuracy across metrics.}
    \label{fig:humanmetric}
\end{figure}

We computed many metrics on the model outputs and generated preferences for each of the pairs that had been annotated. The metrics include PSNR, SSIM, LPIPS, SAMScore \cite{samscore}, and measures of feature similarity from several web-scale models like CLIP \cite{clip}. The level of agreement between each metric and the human preferences across all pairs was computed (i.e., the percentage of pairs where the metric and human pick the same model output as being more similar to the target image). Results are shown in Figure \ref{fig:humanmetric}.

\smallbreak
\noindent \textbf{Finding 1.} \emph{PSNR and SSIM, the most widely used metrics, are insufficient and poorly correlate with human judgments.}
\smallbreak

The correspondences between human preferences and those computed by the standard \sr metrics, PSNR, SSIM, and cPSNR, are very low, as shown in Figure \ref{fig:humanmetric}. In this study, an accuracy of 50\% is equal to random guessing, so metrics with accuracies less than or equal to this are unacceptable for evaluation of \sr quality. This finding aligns with the counter-intuitive values of these metrics in Figure \ref{fig:blur}. Remote sensing \sr should not rely on metrics like PSNR or SSIM when there are other metrics that achieve over 80\% accuracy.

\vspace{-0.1cm}
\subsection{CLIP as an Image Similarity Metric}

Inspired by \metric \cite{clipscore}, we propose a new metric for \sr that measures the distance in image embedding space between the target image and the generated image, using the web-scale CLIP \cite{clip} model. The score is equal to the cosine similarity between the target and output features. This can be done in just a few lines of code, as shown in the snippet below.

\begin{lstlisting}[language=Python]
import clip
import torch.nn.functional as F

# Substitute any CLIP model here
clip_model = clip.load("RN50")

gt_feats = clip_model(gt)
super_res_feats = clip_model(super_res)

score = F.cosine_similarity(gt_feats, super_res_features)

\end{lstlisting}

We assessed several CLIP variants, including CLIPA \cite{clipa}, SigLIP \cite{siglip}, MetaCLIP \cite{metaclip}, and EVA \cite{eva}. Additionally, we trained variations of CLIP from scratch on satellite imagery, similar to \cite{satclip}, but found the performance of these, as metrics, to be worse than the pretrained models. More details are provided in supplementary Section A.2.

\smallbreak
\noindent \textbf{Finding 2.} \emph{CLIP is an effective super-resolution metric.}
\smallbreak
The results from the human-metric correspondence study show that the CLIP models are effective at measuring perceptual quality, with at least 76\% agreement with human preference. CLIPA-v2 \cite{clipav2} performed best in this study, with an impressive agreement accuracy of 84.6\%. We propose this as a measure of image similarity for \sr. 

We use the open-source CLIPA-v2 model with the ViT-bigG-14 architecture, pretrained on DataComp1b \cite{datacomp}, from the open\_clip codebase \cite{openclip}. Note that this model configuration is in the top-5 of best performing models on their set of 38 evaluation tasks including the remote sensing tasks Resisc45 \cite{resisc45} and FMoW \cite{fmow}. The large diversity in the DataComp1b training data is likely a contributing factor to why this model works so well as a metric for satellite imagery.

\vspace{-0.1cm}
\section{Data}
\vspace{-0.1cm}

\begin{table*}
    \centering
    \footnotesize
    \setlength\tabcolsep{5pt}
    \begin{tabular}{|l|a|g|r|a|g|r|a|g|r|a|g|r|}
        \hline
        & \multicolumn{3}{c|}{S2-NAIP} & \multicolumn{3}{c|}{WorldStrat} & \multicolumn{3}{c|}{PROBA-V} & 
        \multicolumn{3}{c|}{OLI2MSI} \\
        \hline
        \cline{2-11}
        Method & cPSNR $\uparrow$ & LPIPS $\uparrow$ & CLIP $\uparrow$ & cPSNR $\uparrow$ & LPIPS $\uparrow$ & CLIP $\uparrow$ & cPSNR $\uparrow$ & LPIPS $\uparrow$ & CLIP $\uparrow$ & cPSNR $\uparrow$ & LPIPS $\uparrow$ & CLIP $\uparrow$ \\
        \hline  
        SRCNN & 17.9039 & 0.3517 & 0.7 & 31.5746 & 0.4493 & 0.6196 & 23.9455 & 0.2112 & 0.7924 
        & 43.4113 & 0.7614 & 0.9285 \\
        HighResNet & 20.9171 & 0.199 & 0.6705 & 32.8787 & 0.4416 & 0.5998 & 22.5687 & 0.2137 & 0.8096 & 
        35.7374 & 0.86 & 0.94 \\
        ESRGAN & 22.6506 & 0.8406 & \textbf{0.8745} & 31.723 & 0.8299 & \textbf{0.9842} & 24.9336 & 0.2154 & \textbf{0.8465} & 36.3202 & 0.8709 & \textbf{0.9518} \\
        SR3 & 19.4706 & 0.6292 & 0.8223 & 30.545 & 0.7745 & 0.9232 & 23.7789 & 0.2115 & 0.8115 & 34.2208 & 0.839 & 0.9391 \\
        \hline
    \end{tabular}
    \caption{Results for four methods on four datasets. The reported LPIPS scores are converted to an accuracy, so with each of these metrics, higher is better. ESRGAN achieves the best \metric for all datasets, and that is the metric we are prioritizing for this study.}
    \vspace{-0.4cm}
\label{tab:methods}
\end{table*}

Existing datasets are relatively small in scale, in terms of number of images and coverage in terms of $km^2$, as reflected in Table \ref{tab:datasets}. Also, most datasets source a portion of their imagery from commercial satellites, making it difficult in practice to extend the data and methods produced.

\subsection{S2-NAIP Dataset}

We build a new dataset, \dataset, consisting of 1.2 million pairs of low-resolution Sentinel-2 time series and high-resolution NAIP images. The goal with this is to measure if an increased scale of training data will improve model performance as well as to release a remote sensing \sr dataset \textbf{built entirely from public domain images}. 

NAIP imagery covers most of the United States with a revisit rate of 2-3 years, and Sentinel-2 is globally available with new imagery every 5-10 days. We source the NAIP imagery from 2019-2020 as well as all spatially overlapping Sentinel-2 images within two months of each NAIP image capture timestamp, resulting in at least 18 Sentinel-2 images. We do not remove images with cloud cover, so that the data maintains a real-world distribution.

\begin{figure*}
    \centering
    \includegraphics[width=\textwidth]{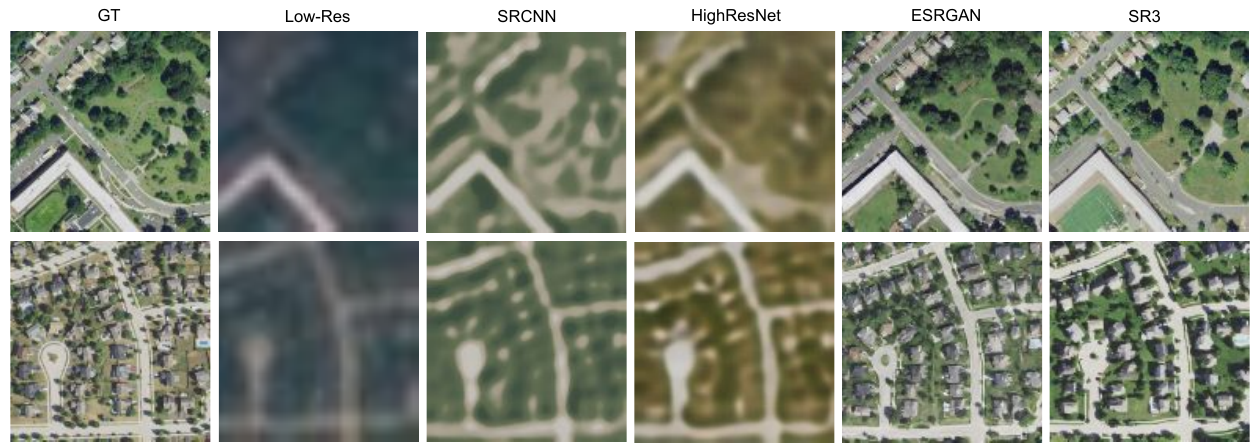}
    \caption{Examples of target images (GT), one of the corresponding low-resolution images (Low-Res), and \sr outputs from SRCNN, HighResNet, Medium ESRGAN, and SR3 on samples from the S2-NAIP dataset. We recommend zooming in for the full effect.}
    \label{fig:model_outputs}
\end{figure*}

The full resolution of each NAIP \textit{tile} is 512x512 pixels. Tiles correspond to Web-Mercator tiles at zoom level 17, i.e., the world is projected to a 2D plane and divided into a $2^{17}$ X $ 2^{17}$ grid, with each tile corresponding to a grid cell. Each NAIP tile corresponds to a time series of Sentinel-2 images, each having a resolution of 32x32 pixels. Geospatial and temporal information for all imagery is provided.

With the goal of avoiding an overload of monotonous landscapes such as ocean or desert, we sample NAIP tiles from within 20km of cities with populations of at least 20k. This resulted in a balance of various man-made infrastructure like water tanks, parking lots, and houses as well as farmland, mountainous area, and forest. 

We train an ESRGAN \cite{esrgan}, as described in Section 5, to upsample the Sentinel-2 images by a factor of four to 128x128 pixels, and use NAIP tiles downsampled by four as the target images. Experiments are run with 1, 3, 10, 30, and 100 percent of the \dataset dataset to determine the difference in performance with varied dataset sizes.

Additionally, we train three sizes of the ESRGAN model on each of these data splits to see if increased model size in conjunction with more data is correlated with better outputs. We define small, medium, and large versions of the ESRGAN with 17mil, 87mil, and 347mil parameters, respectively. Although the effects of data and model size are well studied in the natural image domain, few comprehensive studies have been done on this in remote sensing.

\smallbreak
\noindent \textbf{Finding 3.} \emph{Performance scales with dataset and model size.}
\smallbreak

\begin{figure}
    \centering
    \includegraphics[width=\linewidth]{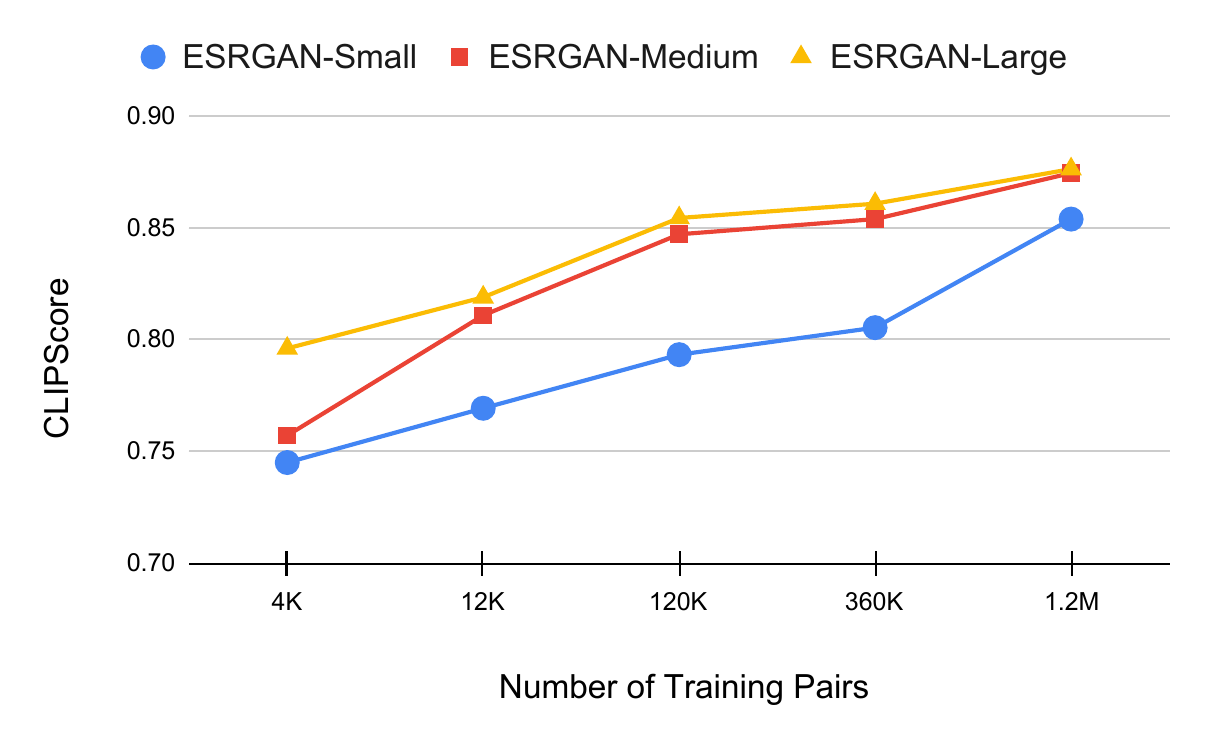}
    \caption{\metric results on the \dataset dataset, with three sizes of the ESRGAN model and five data splits. Performance increases with more data and larger models.}
    \label{fig:sizes}
    \vspace{-0.7cm}
\end{figure}

Somewhat unsurprisingly, we find that training on more of \dataset results in higher performance, as shown in Figure \ref{fig:sizes}. Between the smallest and largest data splits, there is a ten point improvement in \metric for the smallest model. This suggests that training on larger amounts of data will be beneficial to \sr in remote sensing, justifying the creation of large-scale datasets like our proposed \dataset. 

Model size also contributes to higher quality outputs. There is on average a five point improvement between the small and large models across all five data splits. This finding motivates further exploration into larger \sr models for satellite imagery, especially with GANs that provide substantially faster inference than diffusion models.

\vspace{-0.1cm}
\section{Method Study}
\vspace{-0.1cm}


We run a comprehensive study with SRCNN \cite{srcnn}, HighResNet \cite{highresnet}, \cite{worldstrat}, ESRGAN \cite{esrgan}, and Image Super-Resolution via Iterative Refinement (SR3); the latter method is an adaption of Denoising Diffusion for Probablistic Models (DDPM) \cite{ddpm} for \sr. The SR3 model has 97mil parameters so we choose to use the medium ESRGAN from Figure \ref{fig:sizes} which has 87mil parameters, for a relatively fair comparison. Minor model tweaks are made across datasets to account for different upsample factors as well as for stabilization of training. Specific training and model details can be found in supplementary Section A.3.

We train and evaluate these methods on our proposed large-scale \dataset dataset as well as the existing PROBA-V \cite{probav} and the recently proposed benchmarks WorldStrat \cite{worldstrat} and OLI2MSI \cite{oli2msi}. We report \metric as proposed in Section 3 and base our evaluation on this metric, though we also report cPSNR and LPIPS for the sake of comparison to previous work. Results are shown in Table \ref{tab:methods}.

We further experimented with several variations of ESRGAN and describe the findings that led to demonstrably better performance in Section 6. We also ran experiments using Denoising Diffusion Implicit Models (DDIM) \cite{ddim}, which led to much faster inference but slightly lesser quality outputs; as well as Classifier Free Guidance \cite{cfg}, which did not clearly improve output quality. Further details with sample outputs can be found in supplementary Section A.4.

\smallbreak
\noindent \textbf{Finding 4.} \emph{GANs are capable super-resolution methods.}
\smallbreak

Based on the cumulative \metric results on four datasets as shown in Table \ref{tab:methods}, it is clear that ESRGAN is effective in producing high-quality \sr outputs. Compared to training only on L2 loss, like SRCNN and HighResNet, the outputs are sharper and more similar to the target images, as shown in Figure \ref{fig:model_outputs}. 

Although diffusion models have become increasingly popular in computer vision, GANs are proving here to be very effective. It is also important to mention the large gap in inference speed -- GANs require just one forward pass while diffusion models require several. Between our ESRGAN and SR3 models, the ESRGAN is 200x faster. 
\smallbreak
\noindent \textbf{Building Count Study.}
Qualitative assessments of the four models, as shown in Figure \ref{fig:model_outputs}, provides evidence that ESRGAN and SR3 produce high quality outputs. A priority in remote sensing \sr is maintaining accuracy and avoiding hallucinations \cite{probav}. To determine the accuracy of the two generative methods, we run a human study on a held out set of 256 datapoints from \dataset, with the task of counting buildings. We choose this task because human-made structures are of special interest in satellite imagery and are important to represent accurately.

We have an expert annotator count the number of visible buildings in the provided target images as well as the corresponding generated images from ESRGAN and SR3. Of the buildings annotated in each model output, the buildings were labeled as ground truth or hallucination, depending on if the building exists in the target image or not.

Results from this study show that \textbf{ESRGAN accurately generates the buildings from the high-resolution target image 94.70\% of the time}, while SR3 only 81.77\% of the time. Furthermore, the average hallucination rate for each method is 0.41 and 0.43 buildings per image, respectively.

\smallbreak
\noindent \textbf{Finding 5.} \emph{Diffusion models generate realistic but inaccurate images.}
\smallbreak

Results from this study give us a glimpse into the difference in semantic accuracy between two models that both generate high perceptual quality outputs. ESRGAN is more accurate in generating the correct buildings relative to the target image and has a smaller hallucination rate.

\begin{figure}
    \centering
    \includegraphics[width=\linewidth]{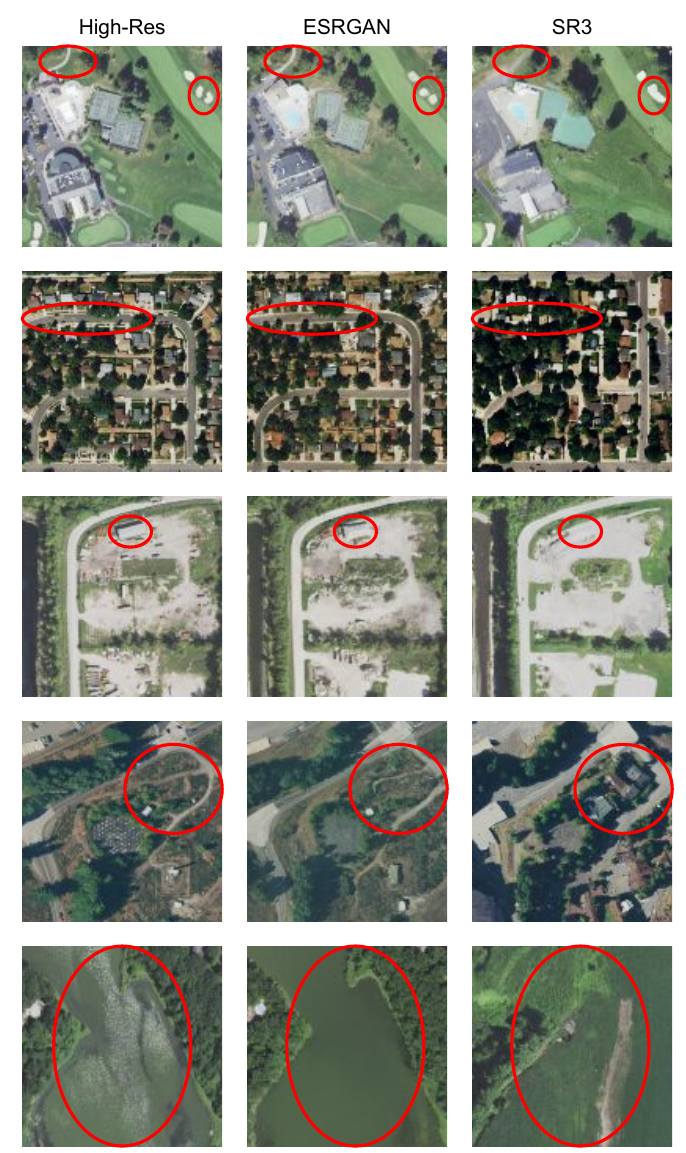}
    \caption{Examples of high-resolution images from S2-NAIP and corresponding ESRGAN and SR3 outputs. Red circles showcase qualitatively incorrect portions generated by SR3.}
    \label{fig:hallucinations}
\end{figure}

Example qualitative outputs from the building count study are shown in Figure \ref{fig:hallucinations}, where red circles point out flaws in the SR3 column such as nonexistent roads, the addition of buildings where there should be none, and a grassy field where there is supposed to be a river. 

\vspace{-0.3cm}
\section{Improving ESRGAN}

We find ESRGAN to be the strongest \sr method of the four compared in our study, and propose novel techniques to improve it. This includes adding an auxiliary loss based on \metric described in Section 6.1 and incorporating domain knowledge into the training pipeline, described in supplementary Section A.5. We hypothesize that these could be transferred to other GAN models.

\subsection{Training with CLIPScore}

Similar to previous works that optimize a loss that is inversely related to a metric such as LPIPS \cite{lpips}, we introduce a \metric loss. We use an L1 loss to minimize the distance between CLIP features of the target and the \sr output. In our experiments, we use CLIP with ResNet50, as it is smaller and thus faster than CLIPA-v2. 

\begin{figure}
    \centering
    \includegraphics[width=\linewidth]{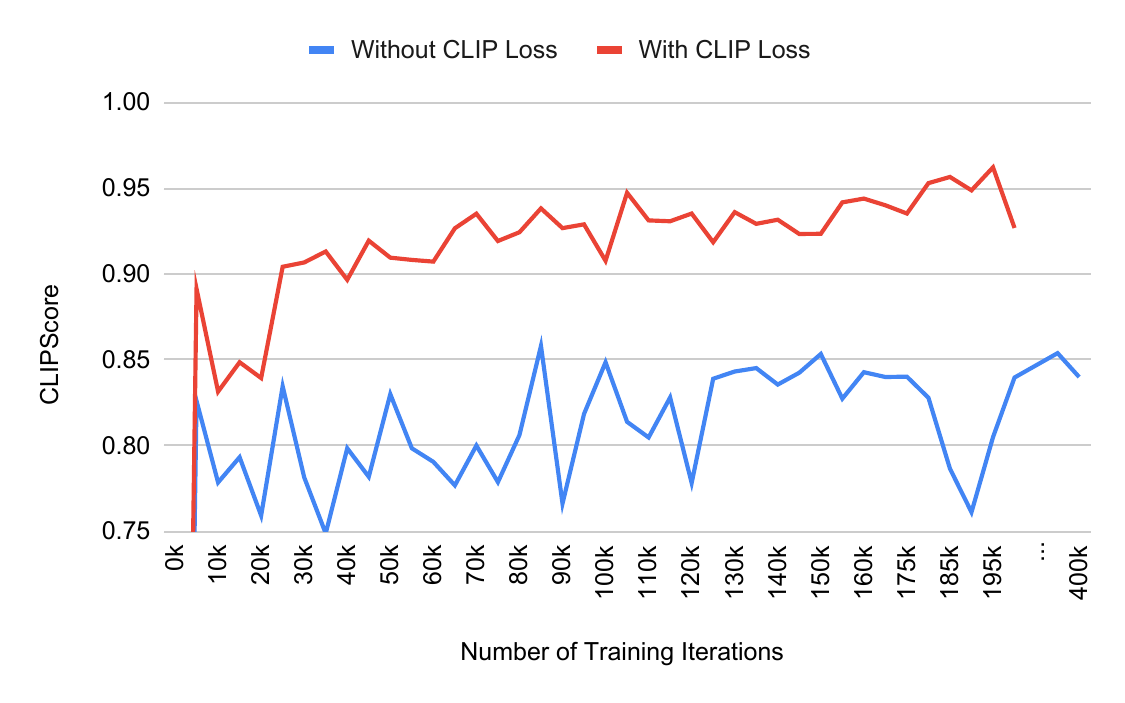}
    \vspace{-0.7cm}
    \caption{Plot of CLIPScore between a model trained with the CLIP-based loss and one without. Note the y-axis is between 0.75 and 0.95 to more clearly show the difference in performance.}
    \label{fig:clipscore}
    \vspace{-0.4cm}
\end{figure}

We train the small ESRGAN in Figure \ref{fig:sizes}, from scratch, with the addition of the  \metric loss. Figure \ref{fig:clipscore} shows results of models trained with and without this loss.

\smallbreak
\noindent \textbf{Finding 6.} \emph{Incorporating CLIP significantly speeds up training and produces more effective models.}
\smallbreak

We find that training a model with the addition of the CLIPScore-based loss results in a \textbf{9 point improvement in CLIPScore in 18x less training time}. This decreases our training time from weeks to hours and the quality of \sr outputs is vastly improved. 
\\

\begin{figure*}
    \centering
    \includegraphics[width=\textwidth]{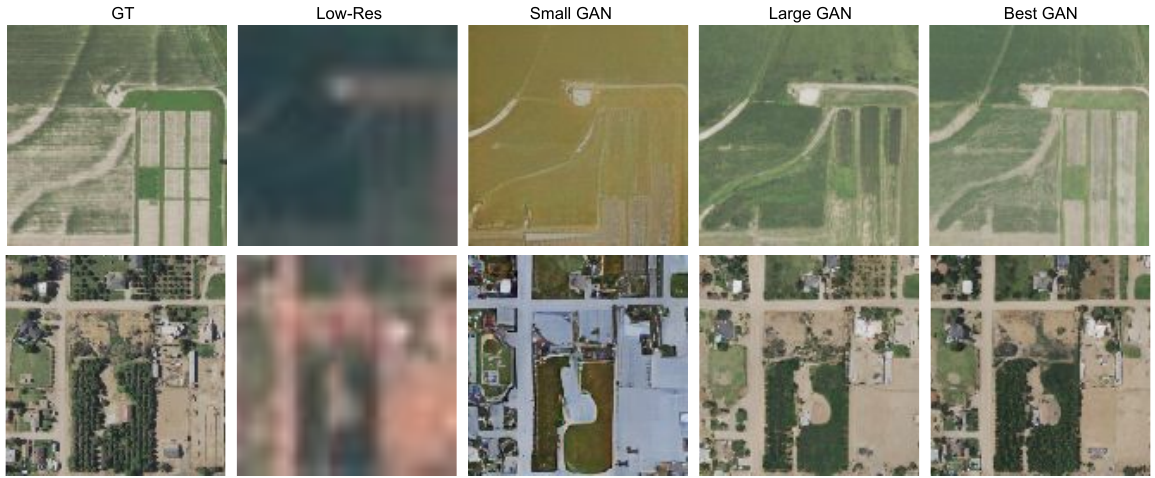}
    \caption{Examples of target images (GT), one of the corresponding low-resolution images (Low-Res), and outputs from the small, large, and best ESRGAN. We recommend zooming in and looking at the fine differences between the last two columns.}
    \label{fig:gan_outputs}
\end{figure*}

\noindent \textbf{The Best Model.}
We compile all of the improvements described in Sections 6.1 and supplementary Section A.5 into one model, and the result is an impressive 0.959 \metric. Figure \ref{fig:gan_outputs} shows qualitative results between the small and large ESRGAN models from Figure \ref{fig:sizes}, compared to the best model. 

\vspace{-0.3cm}
\section{Super-Resolution for Downstream Tasks}

In this section, we analyze two uses of \sr for downstream tasks: images and features. 

We compile a set of five downstream tasks corresponding to existing datasets: the three SatlasPretrain fine-tuning datasets (solar farms, wind turbines, and marine infrastructure) \cite{satlas}, PASTIS (crop type segmentation) \cite{pastis}, and BigEarthNet (land cover segmentation) \cite{bigearthnet}. Because these five datasets are built with Sentinel-2 imagery in mind, most features are relatively coarse, so we also create a new urban land use segmentation dataset from OpenStreetMap labels; details are in supplementary Section A.6.

\subsection{Image Usage}
Previous works such as PROBA-V and WorldStrat \cite{probav, worldstrat} have argued that \sr outputs could be used in place of low-resolution images for downstream tasks like crop type and urban land use segmentation. In fact, the authors of PROBA-V argue that, in remote sensing, \sr methods should be designed for machine rather than human consumption, and suggest that models trained without GAN losses may produce images with more accurate pixel values that are better for machine consumption. 

To evaluate this claim, we generate SRCNN, HighResNet, and ESRGAN \sr outputs from the Sentinel-2 imagery in each of the downstream tasks, and then use them as input to a model with a Swin-Base Transformer \cite{swin} backbone pretrained on ImageNet. Second, we apply a model that inputs eight low-resolution Sentinel-2 images to the same Swin-Base Transformer \cite{swin} backbone pretrained on ImageNet. Results are shown in Table \ref{tab:downstream_images}.

\smallbreak
\noindent \textbf{Finding 7.} \emph{Super-resolution outputs are ineffective for machine consumption.}
\smallbreak

Via this preliminary study we find that using \sr outputs does not outperform inputting the original low-resolution images, as shown in Table \ref{tab:downstream_images}. 
Interestingly, between SRCNN, HighResNet, and ESRGAN, the ESRGAN outputs led to the best average downstream performance. This suggests that optimizing L2 losses does not always imply better outputs for machine use and GANs should be considered for \sr, regardless of whether the aim is machine or human consumption.

\subsection{Feature Usage}

Transfer learning is a very common technique in computer vision for improving the downstream performance, though using \sr weights is largely under-explored. 

We evaluate the effectiveness of this by fine-tuning a Swin-Base Transformer backbone that is (a) randomly initialized; (b) pretrained on ImageNet; (c) pretrained on SatlasPretrain \cite{satlas}; and (d) pretrained on our \dataset \sr dataset. And though not directly comparable, we also apply a similar model with a ResNet50 backbone pretrained using different self-supervised learning methods, including CaCo \cite{caco}, SeCo \cite{seco}, and SSL4EO-S12 \cite{SSL4EO}; Swin weights were not available for these methods. Results are shown in Table \ref{tab:downstream_feats}.

\smallbreak
\noindent \textbf{Finding 8.} \emph{Representations learned through super-resolution transfer to downstream tasks.}
\smallbreak

On average, representations from \sr lead to the best downstream performance compared to those of self-supervised learning methods, CaCo \cite{caco}, SeCo \cite{seco}, and SSL4EO-S12 \cite{SSL4EO}, and from the large-scale SatlasPretrain \cite{satlas} dataset. This encourages further exploration in using \sr representations for downstream tasks.

\begin{table}
    \centering
    \footnotesize
    \setlength\tabcolsep{4pt}
    \begin{tabular}{|l|l|r|r|r|r|r|r|r|}
        \hline
         Input & OSM & S1 & S2 & S3 & Big & PS & \textbf{Avg}   \\
        \hline
        Low-Res & 0.464 & \textbf{0.782} & \textbf{0.888} & \textbf{0.591} & \textbf{0.963} & \textbf{0.348} & \textbf{0.673} \\
         SRCNN & 0.429 & 0.219 & 0.616 & 0.524 & 0.953 & 0.269 & 0.502 \\
         HighResNet & 0.467 & 0.661 & 0.732 & 0.521 & 0.957 & 0.287 & 0.604   \\
         ESRGAN & \textbf{0.486} & 0.723 & 0.706 & 0.524 & 0.954 & 0.299 & 0.615 \\
        \hline
    \end{tabular}
    \caption{Results on 6 tasks (Our OpenStreetMap dataset=OSM, Satlas Tasks=Solar Farms, Wind Turbines, and Marine Infrastructure, BigEarthNet=Big, PASTIS=PS), using an ImageNet Swin Model with 8 low-resolution input images versus \sr outputs from SRCNN, HighResNet, and ESRGAN. }
    \vspace{-0.3cm}
\label{tab:downstream_images}
\end{table}

\begin{table*}
    \centering
    \footnotesize
    \setlength\tabcolsep{4pt}
    \begin{tabular}{|l|l|l|r|r|r|r|r|r|r|}
        \hline
        Arch & Method & OSM & Satlas1 & Satlas2 & Satlas3 & BigEarth & PASTIS & \textbf{Avg}  \\
        \hline
        Res50 & SeCo \cite{seco} & 0.4493 & 0.8468 & 0.8962 & 0.6008 & 
        \textbf{0.9641} & 0.2596 & 0.6695 \\
        & CaCo \cite{caco} & 0.4421 & 0.8384 & 0.8985 & 0.5985 & 0.9629 & 0.2591 & 0.6666  \\
        & SSL4EO \cite{SSL4EO} & 0.4529 & 0.8213 & 0.8924 & \textbf{0.6029} & 0.9637 & 0.2426 & 0.6626  \\
         \hline
        Swin & Random & 0.4446 & 0.7010 & 0.7391 & 0.5591 & 0.9607 & 0.3754 & 0.6299  \\
        & ImageNet & 0.4642 & 0.7820 & 0.8878 & 0.5906 & 0.9631 & 0.3484 & 0.6727 \\
        & Satlas ~\cite{satlas} & 0.4718 & 0.8818 & \textbf{0.9152} & 0.6021 & \textbf{0.9641} & 0.3594 & 0.6991 \\
         & Super-Res & \textbf{0.4759} & \textbf{0.8987} & 0.8958 & 0.5927 & 0.9637 & \textbf{0.3803} & \textbf{0.7012} \\

        \hline
    \end{tabular}
    \caption{Transfer learning results on 6 downstream tasks (Our OpenStreetMap dataset=OSM, BigEarthNet=BigEarth, and the three Satlas tasks are Solar Farms, Wind Turbines, and Marine Infrastructure), comparing several pretraining methods to \sr weights trained on our S2-NAIP dataset.}
\label{tab:downstream_feats}
\end{table*}






\vspace{-0.1cm}
 \section{Deploying Super-Resolution Globally}

With all of these findings, we have deployed global \sr outputs to \url{https://satlas.allen.ai/} for anyone to view. We hope this can further research in AI for remote sensing as well as assist non-ML researchers with annotation tasks such as identifying the drivers of deforestation or counting green energy installations.

\vspace{-0.1cm}
\section{Conclusion}
\vspace{-0.1cm}
We explored, in depth, the metrics, datasets, and methods in the remote sensing \sr field. We propose \metric as a new metric and utilize it to analyze three method types, and find that GANs are very effective for this task. We introduce a new large-scale dataset, \dataset, and determine the benefit of scale. We determine the effectiveness of \sr images and features for downstream tasks. Finally, these findings enable us to train a large-scale model that achieves an impressive CLIPScore.

\appendix

\section{Supplementary Material}

The supplementary material can be accessed at \href{https://pub-25c498004d1e4d4c8da69b2c05676836.r2.dev/Zooming_Out_On_Zooming_In_Supplementary.pdf}{this URL}.

{
    \small
    \bibliographystyle{ieeenat_fullname}
    \bibliography{egbib}
}

\end{document}